# Toward Automated Cognitive Assessment in Parkinson's Disease Using Pretrained Language Models

Varada Khanna*[1], Nilay Bhatt*[1], Ikgyu Shin[1], Sule Tinaz[2#], Yang Ren[1], Hua Xu[1#], Vipina K. Keloth[1$]

**Affiliations:**
[1]Department of Biomedical Informatics and Data Science, Yale School of Medicine, New Haven, CT,
[2]Department of Neurology, Yale School of Medicine, New Haven, CT

* Equal contribution

# Co-senior authors

$ Corresponding author: Vipina K. Keloth (vipina.kuttichikeloth@yale.edu)

**Abstract:**
Understanding how individuals with Parkinson's disease (PD) describe cognitive experiences in their daily lives can offer valuable insights into disease-related cognitive and emotional changes. However, extracting such information from unstructured patient narratives is challenging due to the subtle, overlapping nature of cognitive constructs. This study developed and evaluated natural language processing (NLP) models to automatically identify categories that reflect various cognitive processes from de-identified first-person narratives. Three model families, a *Bio_ClinicalBERT*-based span categorization model for nested entity recognition, a fine-tuned *Meta-Llama-3-8B-Instruct* model using QLoRA for instruction following, and *GPT-4o mini* evaluated under zero- and few-shot settings, were compared on their performance on extracting seven categories. Our findings indicated that model performance varied substantially across categories and model families. The fine-tuned Meta-Llama-3-8B-Instruct achieved the highest overall F1-scores (0.74 micro-average and 0.59 macro-average), particularly excelling in context-dependent categories such as *thought* and *social interaction*. Bio_ClinicalBERT exhibited high precision but low recall and performed comparable to Llama for some category types such as *location* and *time* but failed on other categories such as *thought, emotion* and *social interaction*. Compared to conventional information extraction tasks, this task presents a greater challenge due to the abstract and overlapping nature of narrative accounts of complex cognitive processes. Nonetheless, with continued refinement, these NLP systems hold promise for enabling low-burden, longitudinal monitoring of cognitive function and serving as a valuable complement to formal neuropsychological assessments in PD.

## 1. Introduction

Parkinson's disease (PD) is the second most common neurodegenerative disorder after Alzheimer's disease and currently affects over six million people globally. It is estimated that this number will double by 2040 [1]. Cognitive impairment is very common in PD and associated with more disability, increased caregiver burden, and poor quality of life [2]. There are no effective therapies for cognitive impairment in people with PD. Indeed, the PD community identified maintaining cognitive health as one of its major unmet needs [3]. An in-depth understanding of various cognitive domains (e.g., executive functioning, attention, memory, etc.) that can be impaired in PD will facilitate the development of targeted therapies. Conventionally, these cognitive domains are assessed using standardized neuropsychological tests in controlled clinical or laboratory settings. The assessments are lengthy and licensed examiners who administer the tests are not widely accessible. Moreover, the ecological validity of these tests as it pertains to everyday cognitive skills is limited [4]. In other words, poor performance on standardized tests does not necessarily indicate impaired functioning in everyday life, conversely, intact performance does not guarantee optimal everyday functioning.

In a recent pilot randomized trial (trial ID: NCT05495997) we tested whether personalized mental imagery (MI) training could improve everyday cognition in people with PD [5]. Thirty participants with PD were assigned to either a 6-week MI program, where they practiced vividly imagining everyday tasks, or a control group receiving psychoeducation. In this study, we collected subjective narrative reports from participants with PD, in which they described everyday tasks performed in their naturalistic environments step by step in a detailed narrative. This approach is designed to capture aspects of real-world cognitive functioning that are often not adequately reflected in standardized neuropsychological testing. By encouraging participants to mentally rehearse and recount routine tasks such as preparing a meal, managing medications, or navigating the supermarket, the trial aimed to elicit ecologically valid reflections of cognitive processes spanning multiple domains including sensorimotor, spatiotemporal, executive, social, and emotional. Although not with the same richness as in those narrative reports, these cognitive domains are also reflected in unstructured clinical data, particularly within clinic notes, neuropsychological evaluations, and therapy notes (e.g., speech, occupational, and nursing documentation) of people with PD.

The manual identification and scoring of cognitive domains within these narrative reports by trained raters can be effective, but this procedure is time-consuming, resource-intensive, and subject to inter-rater variability, which poses challenges for scaling the approach to larger cohorts and doing repeated longitudinal assessments that are vital for tracking disease progression and evaluating treatment effects. The manual coding process also limits the granularity with which domain-level impairments can be examined across diverse tasks and patient populations.

Recent advances in natural language processing (NLP), especially the development of pretrained language models (PLMs) including encoder-only models like BERT [6] and decoder-only models like GPT [7] offer a promising solution to these challenges [8–13]. These models have demonstrated their ability in capturing complex language patterns across various domains of knowledge [14,15]. Existing studies using these NLP techniques have mainly focused on extracting/classifying symptoms of PD, classifying patients based on cognition (e.g., normal cognition, PD-MCI, PD-Dementia), and identifying cognitive decline through speech and language (word-finding difficulties) [16–18]. To the best of our knowledge, no prior studies have systematically extracted a broad range of narrative categories reflecting cognitive processes from patient reports. Leveraging the capabilities of PLMs, this study aims to develop and evaluate their performance on extracting seven categories from participant narratives of everyday tasks. The overarching goal is to post-process the extracted information to generate quantitative scores for cognitive domains such as attention, memory, executive function, etc. A longer-term goal is to develop a comprehensive system leveraging PLMs capable of analyzing both speech and clinical text from PD patients, extracting domain-specific cognitive information, and tracking changes longitudinally to monitor disease progression and inform individualized care.

## 2. Background

There have been previous studies conducting linguistic analysis, using language models to detect language patterns in narrative descriptives of people with PD. Yokoi et al. [19] compared spontaneous speech from 53 PD participants and 53 healthy controls using NLP and trained a support vector machine (SVM) model with cross-validation. The model found that PD participants used fewer nouns and more verbs/particles, distinguishing groups with >80% accuracy. Analyzing speech and language from 165 Colombian Spanish speakers, 80 of whom had PD, Escobar-Grisales et al. used advanced models like BERT and Wav2Vec 2.0 for classification. The researchers found that analyzing speech biomarkers was more effective, achieving up to 88% accuracy, compared to analyzing language patterns alone [20].

Researchers have moved beyond simple binary classification to apply linguistic analysis that differentiates between subgroups of people with PD based on their cognitive status. Specifically, Aresta et al. [18] analyzed connected speech to distinguish between PD participants with mild cognitive impairment (PD-MCI), those with normal cognition (PD-nMCI), and healthy controls. Using linguistic features extracted with CLAN (Computerized Language ANalysis) software [21] and classified via SVMs, they achieved up to 85% accuracy in subgroup discrimination. Key markers included retracing, action verb use, utterance errors, and verbless utterances, supporting language as a digital biomarker for early diagnosis and phenotyping. A study conducted by Frattallone-Llado et al. [16] examined these biomarkers across four languages for their potential to track disease-related changes in PD. Their research validated these linguistic markers, finding that PD patients produced shorter narratives, used fewer nouns and auxiliaries, and paused more frequently with reduced pitch variability.

A similar approach was used to monitor changes in patients' condition depending on their medication status in a study conducted by Castelli et al. [22]. They analyzed spontaneous speech from 33 PD patients in ON and OFF medication states. Using machine learning and the Gemma-2

(9B) model [23], they could classify medication states with 98% accuracy and predict neuropsychiatric symptom scores (ρ = 0.81), showing that PLMs can track both motor and psychiatric fluctuations from speech.

Zhang et al. [24] explored large language models for PD treatment planning rather than diagnosis. They integrated patient records and medical guidelines, added retrieval-augmented generation and reasoning steps for transparency, and refined strategies using Monte Carlo Tree Search. Using the Parkinson's Progression Markers Initiative dataset, their system outperformed physician prescriptions and other machine learning methods, reducing symptom severity scores by over 1.4 points on average. A study conducted by Garcia et al. [25] developed an automated language framework to analyze how people retell action- and non-action-based stories, aiming to detect Parkinson's-related semantic impairments from natural speech. They found that reduced use of action-related language reliably identified PD patients, showing the method's potential for scalable, non-invasive diagnosis and cognitive profiling.

While prior studies have primarily focused on linguistic and acoustic biomarkers to classify PD patients, differentiate cognitive subgroups, or track medication states, the present work differs by targeting the extraction of fine-grained categories reflecting various cognitive processes from participant-generated narratives. Instead of using speech or text solely for diagnostic classification, this study leverages PLMs to identify and quantify specific narrative categories expressed in naturalistic descriptions of everyday tasks. This approach has the potential to provide a more interpretable, domain-level assessment of cognition, offering potential utility for longitudinal tracking of cognitive changes in PD. By moving beyond binary classification and toward structured representation of the elements of cognitive processes, our work establishes a framework for developing systems that can process both clinical text and speech to monitor cognitive function dynamically over the course of disease progression.

## 3. Methods

A schematic overview of the entire study process is shown in Figure 1.

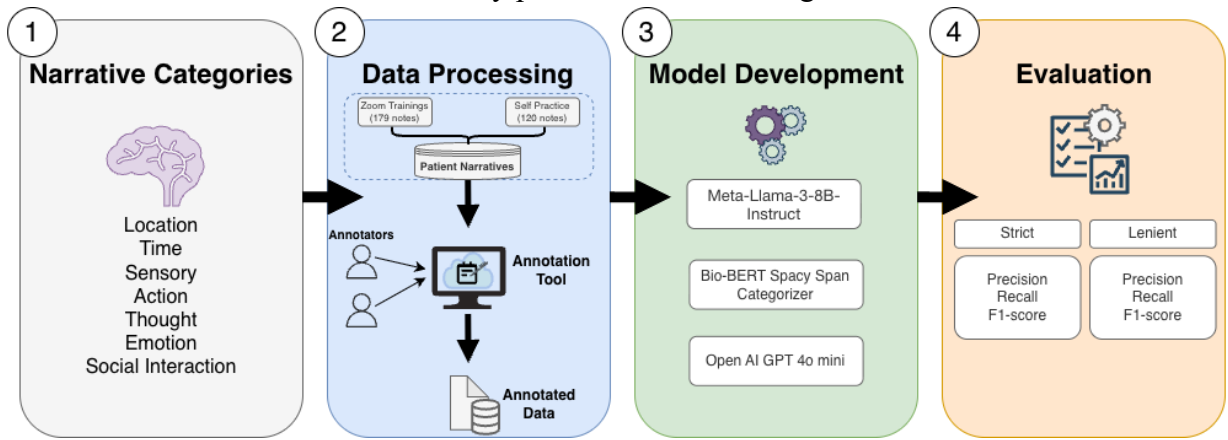

**Figure 1.** A schematic representation of the overall workflow.

## 3.1 Dataset and Annotation

Narrative reports were collected from participants as part of a pilot clinical trial (trial ID: NCT05495997), which included people with a diagnosis of PD according to the Movement Disorders Society (MDS) diagnostic criteria [26] who were 40 years of age or older. A total of 30 participants were randomly assigned to two groups - the experimental PD-MI group that received mental imagery (MI) training and the control PD-Con group that received psychoeducation on cognitive difficulties experienced in PD for four weeks. The data selected for this study is from the PD-MI group (15 participants) and entails transcripts of 179 one-on-one zoom training sessions with a research team member (each of the 15 participants having 12 trainings, one training redacted) and 120 self-practice sessions (15 participants, 8 sessions) that the participants completed in-between zoom training sessions. The MI training was designed to improve goal-directed, everyday cognitive functioning of people with PD, which relies on multiple cognitive domains including executive function, episodic memory, integration of multisensory input with action, social interaction, and emotion regulation. Four categories of MI training tasks were created to capture the variety of everyday tasks: household tasks (e.g., preparing a meal, organizing a closet), outdoor activities (e.g., going to the grocery store, visiting the post office, yard work), event planning (e.g., arranging a vacation, party, or charity event), and social engagement (e.g., celebrating with family or dining with friends). Participants were instructed to imagine/mentally simulate the implementation of these tasks from a first-person perspective. They were trained to include contextual cues (when, where, what, who) to induce episodic specificity in their imagery and to integrate multiple sensations (visual, auditory, tactile, body sensations, etc.) as well as actions, thoughts, and feelings to facilitate imagery vividness. Participants chose one task per session, outlined the main steps and then elaborated on all these features associated with the task to create the imagery narrative reports.

We extracted the following narrative categories from these reports: Location, time, sensory, action, thought, emotion, and social interaction. Table 1 presents the categories together with their descriptions and illustrative examples.

**Table 1.** Categories with descriptions and examples

| Category | Description | Examples |
| --- | --- | --- |
| Location | References to places or settings where actions or events occur. It includes specific locations (e.g., a room or city) or general descriptions of spatial positioning. | "Forest," "Backyard," "I-95 highway," "Paris." |
| Time | This category captures temporality, as well as references to time, including specific | "Monday," "April 2024," "Middle of the week." |

|  | times of day, days of the week, months, seasons, and holidays or special days. |  |
|---|---|---|
| Sensory | This category involves the senses of sight, hearing, smell, taste, and touch (including temperature, pressure, and pain), as well as awareness of body position and internal physiological states like hunger, thirst, or heartbeat. | "The sky was bright and clear," "I felt a sharp pain in my leg," "Birds chirping,", "Sweet tea," "Nice fragrance" |
| Action | Descriptions of purposeful physical or mental actions performed by the participant or others. | "The patient walked to the bathroom," "I filled out a form." |
| Thought | Captures internal mental processes and reflections that do not necessarily lead to concrete actions or plans. It reflects how the participant interprets, recalls, or mentally processes experiences and observations. | "I decided to leave early," "I realized I had made a mistake." |
| Emotion | Participants' emotional experiences or feelings. It includes both positive and negative emotions. | "It was a happy time," "The movie made me nostalgic." |
| Social interaction | This category includes references to interactions or exchanges between people. It covers both verbal communication and non-verbal social interactions, such as gestures or expressions. | "I spoke with the doctor," "We laughed together." |

All narrative reports were de-identified using Robust-DeID [27], which systematically replaced protected health information (PHI) with category-specific placeholders (e.g., [PERSON], [STREET]). Following de-identification, the reports were manually annotated for the seven categories. Annotation guidelines were developed to ensure consistency, containing detailed instructions and definitions for each category. The full guidelines are provided in the Supplementary Materials. Three annotators underwent multiple rounds of structured training, and each report was independently annotated by two annotators. To assess reliability, inter-annotator agreement was calculated after each round. Discrepancies were resolved through adjudication with expert input (V.K.K., S.T.). In the final round, Cohen's Kappa scores reached 0.79, 0.7, and 0.85 across annotator pairs, indicating substantial agreement. Similarly, the average entity F1 scores for annotator pairings were 0.91, 0.86, and 0.93, confirming the robustness of the annotation process.

The distribution of the seven categories in the annotated dataset (Fig. 2) was highly imbalanced, with certain domains (e.g., *action* and *location*) appearing much more frequently than others, such as *emotion*. This imbalance reflects the natural variability in how participants describe everyday cognitive processes in their narratives. To ensure that all categories were adequately represented during model training and evaluation, we performed a stratified split of the data, maintaining proportional representation of each category across the training (70%), development (10%), and test (20%) sets.

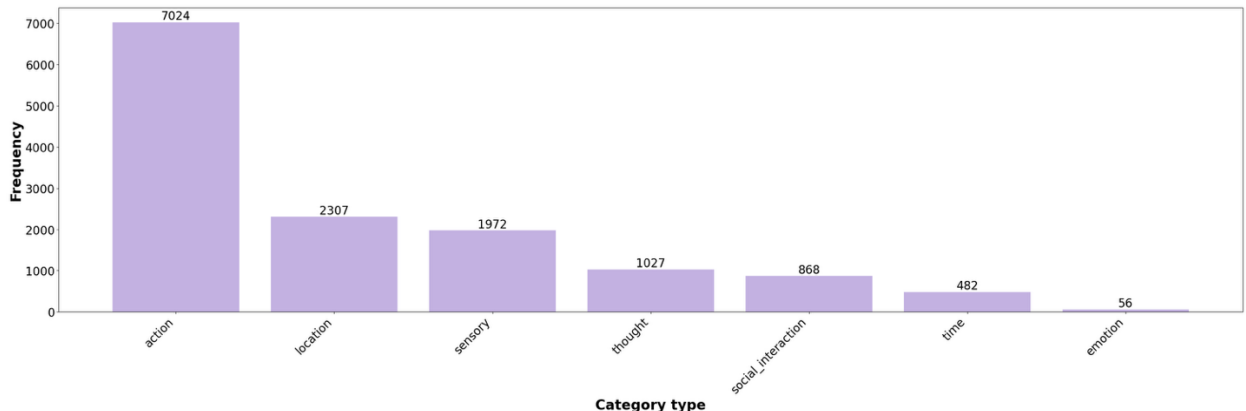

**Figure 2.** Frequency distribution of the seven categories in our annotated dataset.

### 3.2 Model development
We developed and evaluated a series of models to extract categories from participant narratives, ranging from specialized transformer-based models to instruction-tuned language models. Specifically, we trained a Bio_ClinicalBERT span categorization model for structured entity extraction, fine-tuned a Meta-Llama-3-Instruct 8B model using a QLoRA setup for domain-specific instruction following, and evaluated GPT-4o mini in zero- and few-shot settings to assess generalization without task-specific fine-tuning.

**Bio_ClinicalBERT:** We developed a model to automatically identify category mentions from participant narratives using the spaCy framework with a Bio_ClinicalBERT transformer backbone [28]. We employed the SpanCategorizer architecture because the task involved nested named entity recognition (NER), where multiple category mentions can overlap or be embedded within the same text span. Traditional NER models are designed for sequential tagging of tokens and struggle to correctly capture nested structures. For example, in the sentence "She felt pain while opening the hot oven," the span "felt pain" is tagged as *sensory*, while the phrase "opening the hot oven" is an example of overlapping entities with the whole phrase tagged as *action*, and "hot oven" tagged as *sensory*. Such overlap cannot be captured by sequence tagging approach, which assigns a single label per token. The span-based approach in spaCy allows independent classification of candidate text spans, making it better suited for this type of fine-grained, multidomain cognitive annotation compared to standard token classification methods. The model was designed to detect text spans corresponding to the seven categories defined in our annotation schema.
Hyperparameter tuning was performed with learning rates of 2e-5, 3e-5, and 5e-5. The model was evaluated after each epoch on the development set using span-level F1 as the primary metric, with per-epoch checkpoints saved and the best-performing model retained. The best performance on the development set was obtained with a learning rate of 3e-5, a batch size of 4, a dropout rate of 0.1, and Adam optimization using spaCy's default learning schedule.

**Meta-Llama-3-8b-Instruct:** We fine-tuned the Meta-Llama-3-Instruct 8B model for category-specific instruction following using a QLoRA [29] setup to efficiently adapt the model while reducing memory requirements. The training data consisted of annotated participant narratives reformatted into an instruction–response demonstration, where the *instruction* described the annotation task, guidelines and definitions of categories, the *input* contained sentences from the participant narratives, and the *response* was the corresponding gold-standard annotation in structured JSON format (see *Supplementary Materials* for the detailed prompt). This format enables the model to follow category-specific annotation rules and produce output in a standardized schema. To enable efficient fine-tuning, the base model was quantized to 4-bit precision and configured with low-rank adaptation (LoRA) layers (rank = 16, $\alpha$ = 64, dropout = 0.05). At inference, we used the vLLM package for efficient and reproducible text generation. A temperature of 0 was used to enforce deterministic outputs for consistent evaluation. The model predicted category type and corresponding text spans.

**GPT-4o mini:** GPT-4o mini was used under zero-shot and few-shot learning settings to extract the seven predefined categories from participant narratives. In the zero-shot setting, the model was provided only with general task instructions and definitions of the categories along with instructions on formatting the output response. In the few-shot setting, five representative examples collectively covering all seven categories were included in the prompt to guide the extraction process. The model was then evaluated on the test set to assess its ability to identify categories under both conditions. The full prompt templates, including instructions, definitions, and examples, are described in the *Supplementary Materials*.

### 3.3 Evaluation

Model performance was evaluated using precision, recall, and F1-score, computed under both micro and macro averaging schemes to capture overall and per-class performance, respectively. To account for variability in span boundary alignment, we applied two evaluation criteria: *strict and lenient*. Under *strict* criterion, a prediction was considered correct only if both the category type and the exact text span boundaries matched the gold-standard annotation. In contrast, the lenient criterion allowed for partial overlap between predicted and true spans, as long as the category type was correctly identified.

### 4. Results

The results (Fig. 3) show consistent trends across both micro- and macro-averaged evaluations. Under strict matching criteria (Fig. 3A and 3C), Llama-3-8b-Instruct achieved the highest overall performance, with precision, recall, and F1-scores closely aligned around 0.7 for micro-averages and around 0.55 for macro-averages. Bio_ClinicalBERT showed slightly higher precision, but lower recall compared to Llama-3-8b-Instruct. In contrast, GPT-4o mini models performed substantially lower under zero-shot settings, though performance improved under five-shot prompting, particularly in recall and F1-scores. The lenient evaluation (Fig. 3A and 3C) led to uniformly higher scores across all models, with the greatest gains observed for GPT-4o mini (five-shot), indicating improved recognition of partially correct predictions under relaxed criteria. These results demonstrate that Llama-3-8b-Instruct yields higher precision, recall, and F1 than other models across both micro- and macro-averages, with the performance gap most pronounced for recall and F1.

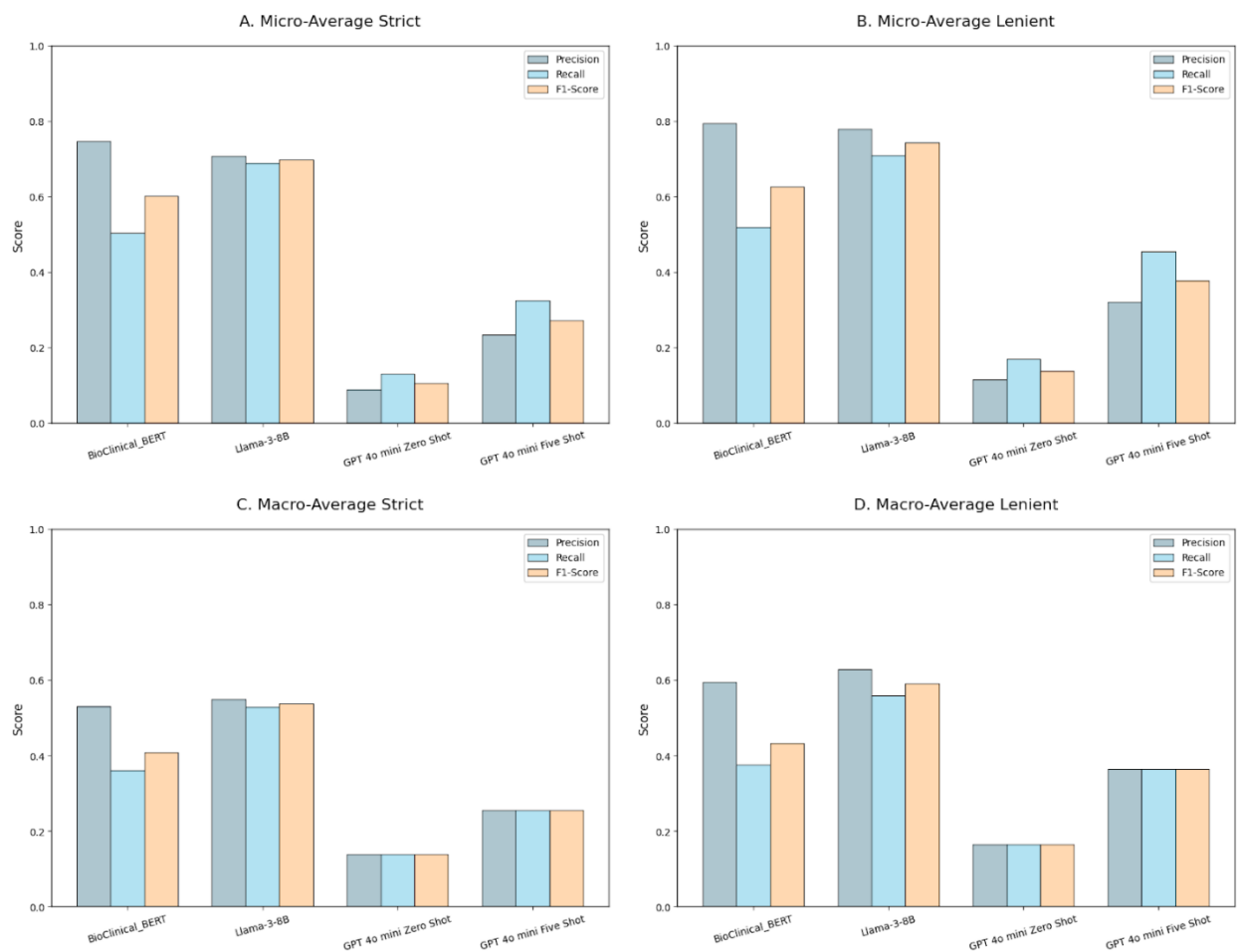

**Figure 3.** Overall model performance across evaluation settings.

The entity-level comparison of F1-scores (Fig. 4A and 4B) highlights notable variability across categories and models. Under strict evaluation (Fig. 4A), Llama-3-8b-Instruct achieved the highest F1-scores across most categories, particularly for *thought* and *social_interaction*, indicating strong generalization in complex categories. Bio_ClinicalBERT performed comparably for *sensory* and *location* and improved performance for *time*, but showed lower performance for categories such as *thought, social_interaction* and *emotion*. Both GPT-4o mini zero-shot and five-shot models exhibited relatively low F1-scores across categories, though five-shot prompting improved performance in *thought* and *social_interaction*. Performance in the *emotion* category was relatively low for both Llama-3-8b-Instruct and Bio_ClinicalBERT, likely reflecting the limited number of available samples (n=56) in this category. The small sample size may have constrained the models' ability to generalize and accurately identify emotion-related entities. In contrast, GPT-4o mini demonstrated comparatively better performance for this category. Under the lenient evaluation (Fig. 4B), scores increased across all models and categories, with the largest gains for GPT-4o mini five-shot. *Supplementary Table T1* presents precision, recall, and F1-scores for all seven categories across all evaluated models.

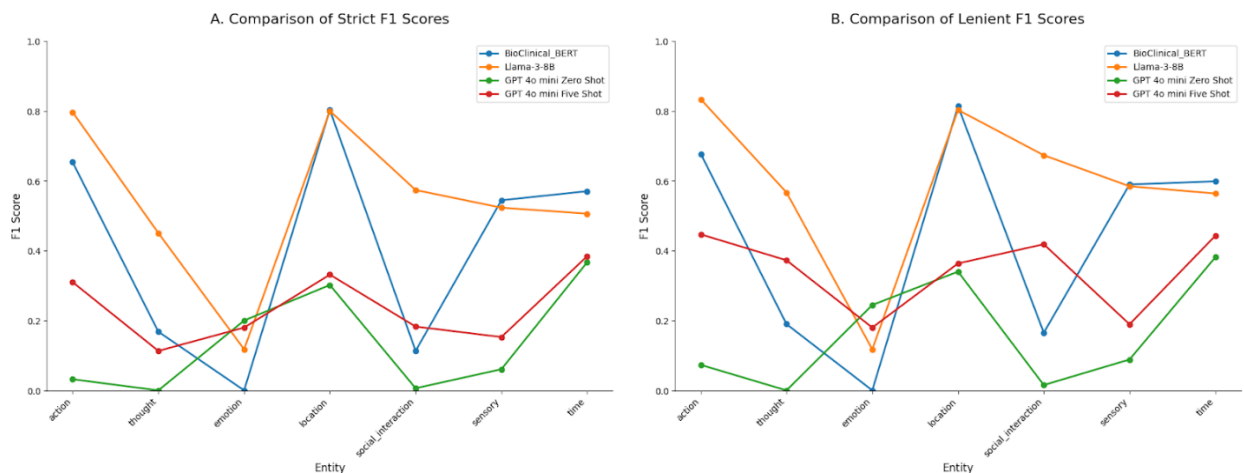

**Figure 4.** Category-level comparison of model F1-scores.

## 5. Discussion

This study demonstrates the feasibility and limitations of using large LLMs and domain-specific transformer architectures to automatically extract narrative categories from naturalistic reports of goal-based everyday tasks obtained from individuals with PD. Unlike prior work that has largely focused on identifying linguistic or acoustic biomarkers for PD diagnosis or subgroup classification, our study aims to extract distinct categories that serve as the building blocks of cognitive processes employed during everyday activities from participant-generated descriptions. This approach addresses limitations of traditional standardized neuropsychological tests by leveraging ecologically valid first-person narratives that better reflect everyday cognitive functioning.

The performance comparison across models reveals clear differences in both overall accuracy and category-specific sensitivity. The per-category breakdown provides insight into which narrative categories are easier or more difficult for models to recognize. Llama-3-8b-Instruct outperformed Bio_ClinicalBERT and GPT-4o mini, particularly in recognizing complex categories such as *thought* and *social_interaction*. This demonstrated that fine-tuning results in better generalization to these harder categories, capturing more subtle contextual signals, suggesting that larger language models may be better suited for extracting cognitively complex and less overt phenomena. While Bio_ClinicalBERT achieved strong performance in well-defined categories such as *time* and *location*, it struggled with abstract categories that require contextual inference. The near zero performance by Bio_ClinicalBERT on *emotion* aligns with the category distribution (Fig. 1), where emotion is underrepresented (n=56).

From a resource utilization perspective, BERT models remain the most efficient, with faster inference and lower computational requirements, making it suitable for large-scale or real-time applications. However, these advantages come at the expense of lower adaptability to complex, nested entity relationships inherent in narrative data. Llama models, though more resource-intensive, demonstrate superior capacity to model such instances, handling the nested NER structure of narrative category extraction more effectively. These models better capture subtle contextual dependencies where multiple overlapping cognitive processes are expressed within a single sentence and can learn better from fewer samples compared to BERT models.

**Qualitative analysis of model shortcomings**: To perform a qualitative error analysis, we manually examined the model outputs for 50 randomly selected sentences from the test set.
Our qualitative error analysis revealed that the Bio_ClinicalBERT with the spaCy Span Categorizer exhibited a strong bias toward under-tagging. In 22 out of 50 sentences, no spans were detected, and only 40 annotations were produced overall. Spans were short and literal, typically one or two words, indicating reliance on explicit lexical cues rather than contextual meaning. The model recognized categories only when clearly signaled by surface features such as concrete verbs or place names. *Action* tags were mostly accurate for simple verbs (e.g., "walked," "examined") but missed complex or multi-step activities. *Thought* and *social_interaction* tags had a higher precision but a much lower recall as the model seemed to struggle with abstraction unless explicit verbs related to those categories were mentioned. As reflected in the results *emotion* tags were entirely absent. *Sensory* tagging was inconsistent, sometimes identifying perceptual details but most times mislabeling unrelated items. *Location* was the only consistently accurate category, with clear span boundaries and correct identification of spatial expressions. Overall, the model exhibited high precision but very low recall, missing many of the cognitive domain mentions.

For the fine-tuned Llama model, *action* tags were the most frequent and stable but often merged multiple events into a single span (e.g., tagging *"went to the store and bought groceries"* as one event while our annotations split them into two events). *Thought* tagging occurred only when explicit cognitive verbs appeared (e.g., "I think…," "he planned…"). *Emotion* was rare and appeared only when direct affective terms were used (e.g., "happy," "angry"), showing little inferential understanding of emotional context. *Social_interaction* tagging captured interpersonal references but frequently overextended to general group mentions (e.g., labeling "the team" or "the crowd" without interaction). *Sensory* tagging was inconsistent, often applied to tangible nouns or quantities instead of perceptual experiences. In contrast, *location* and *time* tagging were the most reliable and contextually accurate, correctly identifying spatial and temporal expressions (e.g., "his kitchen," "Monday") with minimal fragmentation. Overall, the model performed consistently in terms of not having a strong bias towards either making too many false alarms or missing too many actual positive cases.

For GPT-4o mini, the core issues for the model's zero-shot performance manifested as null responses and span and boundary errors, with the model frequently tagging individual words rather than complete phrases, leading to mistagging and overtagging. Category-specific performance was inconsistent: *action* and *location* were comparatively well-captured, though a*ction* over-generalized by tagging all verbs (including non-patient actions), and *location* misinterpreted the definition by tagging all directional terms (e.g., "right," "left"). *Social_interaction* and *time* suffered from fragmentation; individual pronouns (e.g., "she," "he") and nouns (e.g., "Niece") were erroneously tagged for *social_interaction*, while for *time* phrases were split (e.g., "first" and "Birthday"). The context-dependent categories of *thought* and *emotion* were also missed, capturing only isolated single words and failing to identify relevant phrases. For the five-shot analysis, the model showed clear improvements. There were no null outputs, but the challenge of "literal extractions," where individual words were tagged instead of full phrases, continued. The model extracted categories like *action*, *location*, *time*, and *social_interaction* better, but this sometimes led to over-tagging or replacing one category type with another, especially when contextual understanding was needed. Category-specific performance improved but retained key issues. *Action* was mostly correct, but boundary issues and the tagging of observations (non-first-person actions) persisted. *Emotion* was able to capture all instances but still had span boundary issues. *Social_interaction* was captured much better, but the model still incorrectly tagged individual pronouns and mentions of people. *Sensory* was sometimes confused with *social_interaction* and the model still struggled with *location* definitions. *Thought* tagging remained challenging; even with examples provided it had boundary issues, missed tagging, and was often replaced by *sensory* tags. Providing explicit instructions, detailed definitions and more examples as part of the prompt might improve performance of GPT models for specific error patterns observed.

This study has several limitations. First, the dataset size was modest and derived from a single pilot trial, which may limit model generalizability. Future work should expand training data to include a broader range of participants and narrative contexts, potentially integrating data from EHRs and speech transcripts. Second, the study focused only on seven narrative categories; however, cognition is a broad and complex construct that encompasses additional domains not specifically addressed here, such as executive function, memory, etc. Some of these domains often manifest in more complex or implicit ways in natural language, making them more difficult to capture through direct annotation. Future work will need to expand the annotation schema to incorporate these domains, potentially by integrating hierarchical or multi-level annotation frameworks and leveraging LLMs for guided weak supervision. Third, our experiments were limited to a small set of models and configurations. While these provide valuable insights into the feasibility of automated narrative category extraction, they do not represent the full landscape of available LLMs. Newer models such as GPT-5 or GPT reasoning series, or larger-parameter variants of LLaMA may offer improved performance, particularly in handling nuanced or overlapping categories. Future research will explore these newer architectures and optimization strategies to assess their potential for more accurate and context-aware extraction. Additionally, the study's reliance on human annotation introduces an element of subjectivity, even with rigorous guidelines and high inter-annotator agreement. Cognitive constructs are inherently abstract, and subtle differences in interpretation among annotators can influence the quality and consistency of the labeled data. These variations may, in turn, impact model learning and downstream performance.

## 6. Conclusion

This work establishes a foundation for computational extraction of categories that reflect cognitive processes using LLMs. By automating the identification of these categories from first-person narratives, such systems could eventually provide continuous, low-burden cognitive monitoring and complement formal neuropsychological testing. When integrated into longitudinal assessments or clinical documentation, these tools may support earlier detection of cognitive decline, personalized therapy design, and more responsive care for individuals with PD.